\documentclass[sigconf,nonacm]{acmart}
\usepackage{multirow}
\usepackage{colortbl}

\AtBeginDocument{%
  }

\setcopyright{acmlicensed}
\copyrightyear{2025}
\acmYear{2025}
\acmDOI{XXXXXXX.XXXXXXX}

\acmConference[ICKG '2025]{The 16th IEEE International Conference on Knowledge Graphs}{Nov 13--14,
  2025}{Limassol, Cyprus}
\acmISBN{978-1-4503-XXXX-X/18/06}

\begin{document}

\title[Generative and Contrastive Graph Representation Learning]{Generative and Contrastive Graph Representation Learning}

\author{Jiali Chen}
\authornote{Jiali Chen is the corresponding author.}
\email{jiali_chen2@apple.com}
\affiliation{%
  \institution{Apple}
  \city{Cupertino}
  \state{California}
  \country{USA}
}

\author{Avijit Mukherjee}
\email{avijit_mukherjee@apple.com}
\affiliation{%
  \institution{Apple}
  \city{Cupertino}
  \state{California}
  \country{USA}
}

\renewcommand{\shortauthors}{Chen et al.}

\begin{abstract}
Self-supervised learning (SSL) on graphs generates node and graph representations (i.e., embeddings) that can be used for downstream tasks such as node classification, node clustering, and link prediction. Graph SSL is particularly useful in scenarios with limited or no labeled data. Existing SSL methods predominantly follow contrastive or generative paradigms, each excelling in different tasks: contrastive methods typically perform well on classification tasks, while generative methods often excel in link prediction. In this paper, we present a novel architecture for graph SSL that integrates the strengths of both approaches. Our framework introduces community-aware node-level contrastive learning, providing more robust and effective positive and negative node pairs generation, alongside graph-level contrastive learning to capture global semantic information. Additionally, we employ a comprehensive augmentation strategy that combines feature masking, node perturbation, and edge perturbation, enabling robust and diverse representation learning. By incorporating these enhancements, our model achieves superior performance across multiple tasks, including node classification, clustering, and link prediction. Evaluations on open benchmark datasets demonstrate that our model outperforms state-of-the-art methods, achieving a performance lift of 0.23\%-2.01\% depending on the task and dataset.
\end{abstract}

\ccsdesc[500]{Computing methodologies~Artificial intelligence}
\ccsdesc[500]{Computing methodologies~Machine learning}
\keywords{Graph Representation Learning,  Self Supervised Learning, Graph Neural Network, Contrastive Learning, Generative Learning}


\graphicspath{{./images/}}

\maketitle


\section{Introduction}
\label{introduction}
Self Supervised Learning (SSL) on graphs is an active area of research~\citep{graphmae2022,mvgrl2020,bgrl2022,dgi2019,gcc2020,graphsage2017,you2020}. Graph SSL methods are particularly useful in cases where either there are no labels to train supervised models or the availability of labels is sparse~\citep{kipf2017}. Using SSL on graphs, one can generate node embeddings, which can be further applied to downstream tasks such as node and graph classification and link prediction.

There are two predominant methods for graph SSL that have been studied and compared. While \textit{contrastive learning} methods have been the state-of-the-art for node and graph classification tasks~\citep{mvgrl2020,gcc2020,bgrl2022,you2020,zhu2020a,zhu2020b}, \textit{generative} methods, such as graph auto-encoder (GAE), have gained momentum in the recent years, particularly for link prediction tasks~\citep{graphmae2022,maskgae2023,cui2020adaptive,hu2020,pan2018,kipf2016}. Contrastive learning in graph SSL works by generating augmented views of the input graph and pulling the node embeddings of positive pairs closer while pushing apart negative pairs by the construct of the loss function. For large graphs, this can be computationally expensive as there is a need to construct a large number of postive and negative pairs for each node in the graph. Also, there is no principled way to generate postive node pairs. Random walk~\citep{perozzi2014deepwalk} and neighborhood sampling ~\citep{graphsage2017} are two common choices but, they fail to identify sub-graphs that are far apart and have similar local structure. While a recent paper, \textit{BGRL}~\citep{bgrl2022}, proposed a way to avoid the computational burden, the strong reliance of contrastive methods on augmentation strategies still remains open. Another notable limitation of contrastive methods, as mentioned in~\citep{s2gae2023}, is that they do not perform well on link prediction tasks.

GAE methods~\citep{cui2020adaptive,hu2020,pan2018,kipf2016} work by first passing the input graph through an encoder to generate embeddings of nodes, followed by a decoder that attempts to reconstruct the input graph structure and features. One primary advantage of this approach is avoiding some of the aforementioned challenges faced by constrastive learning methods. By construct, GAEs generally work well in link prediction tasks but underperform in node and graph classification tasks when compared to constrastive methods. Recent methods~\citep{graphmae2022,s2gae2023,maskgae2023,shi2023gigamae} have overcome some of these limitations by masking parts of the input graph, such as node features and edges, and by attempting to predict those following the decoder step. The masked auto-encoder based methods are motivated by the success in language models~\citep{bert2019} and computer vision~\citep{he2021}. However, to date, none of the GAE methods have outperformed contrastive methods comprehensively in all three tasks: node classification, node clustering, and link prediction.

\subsection{Related Literature}
\label{litreview}
In the remainder of this section, we review prior research on graph self-supervised learning (SSL) that is most relevant to our work, followed by a discussion of our contributions. Later, in the experimental section, we evaluate the performance of our model on benchmark datasets and compare it against state-of-the-art methods across different categories. Among contrastive graph SSL methods, we include \textit{BGRL}~\citep{bgrl2022} and \textit{MVGRL}~\citep{mvgrl2020}. For generative graph autoencoder (GAE)-based approaches, we consider \textit{GraphMAE}~\citep{graphmae2022}, \textit{S2GAE}~\citep{s2gae2023}, \textit{GiGaMAE}~\citep{shi2023gigamae}, and \textit{MaskGAE}~\citep{maskgae2023}. Additionally, we compare our model with \textit{GCMAE}~\citep{wang2024generative}, a hybrid approach that integrates both contrastive and generative objectives.

Contrastive learning methods for graph SSL rely heavily on techniques to generate augmented views of the input graph. Some of the common techniques to generate augemented views are: edge perturbation~\citep{hu2020,you2020,zeng2021}, diffusion~\citep{mvgrl2020,zekarias2021}, node feature masking~\citep{zhu2020b}, and neighborhood sampling~\citep{perozzi2014deepwalk,graphsage2017}. Another key requirement of contrastive methods is negative sampling, which amplifies the computational burden. In \textit{MVGRL}, diffusion is used to generate augmented views of the input graph. The input graph and the augmented views are sub-sampled and fed to two dedicated graph encoders followed by a shared Multi-Layer Perceptron (MLP) to learn node embeddings. While any graph neural network can be used for the encoders, the authors chose Graph Convolutional Network (GCN)~\citep{kipf2017} in their experiments. The learned embeddings are then fed to a graph pooling layer followed by a shared MLP to learn graph embeddings. The loss function attempts to maximize the Mutual Information (MI) between two views by contrasting node embeddings of one view with graph representation of the other view and vice versa. 

\textit{BGRL}~\citep{bgrl2022} proposed a constrastive graph SSL while circumventing negative sample generation. In this method, node embeddings are computed by encoding two augmented versions of a graph using two distinct graph encoders~\citep{kipf2017}: an online encoder, and a target encoder. The online encoder is trained through predicting the representation of the target encoder, while the target encoder is updated as an exponential moving average of the online network. By avoiding the negative sample generation step, BGRL can scale to very large graphs.

To introduce graph-wise information to contrastive learning, \textit{AMSGCL}~\citep{fu2024automated} employs a multi-scale contrastive framework designed to capture both node-level and graph-level semantics in an unsupervised learning setting. Traditional contrastive methods often rely on a single scale, such as node-to-node or node-to-global contrasts, which can limit the model's ability to learn rich, multi-level information from the graph. AMSGCL overcomes this limitation by combining node-to-node and node-to-global contrastive methods, thus enabling the encoder to simultaneously learn from various structural levels within the graph. Furthermore, AMSGCL introduces an automated weight search mechanism guided by homophily, which dynamically assigns optimal weights to each contrastive loss scale based on dataset-specific characteristics. This weight optimization enhances the model’s adaptability and overall performance across different datasets. Experimental results demonstrate that AMSGCL achieves superior results in node classification, highlighting the effectiveness of incorporating graph-level information through multi-scale contrastive learning.

In \textit{MaskGAE}~\citep{maskgae2023}, a subset of edges in the input graph are masked, followed by an encoder which generates node embeddings. Thereafter, two decoders, one aimed to reconstruct edges while another aimed to recover the degree of nodes is applied. The loss function is a combination of edge reconstruction loss and node degree prediction error. MaskGAE achieved comparable results on node classificaiton tasks as that of GraphMAE but it outperformed the latter on link prediction tasks on benchmark data. \textit{S2GAE}~\citep{s2gae2023} applies edge masking in a similar manner as MaskGAE, however, it differs in the decoder step. The cross-correlation decoder in S2GAE computes correlation between node embeddings computed in each layer of the encoder, which is typically a GCN~\citep{kipf2017}. This is followed by a MLP layer which attempts to predict the existence of an edge in the input graph -- i.e., edge reconstruction being the objective. Based on the experimental results, S2GAE outperforms both constrastive as well as prior GAE methods in both node classification and link prediction tasks, particularly when applied to large datasets.

\textit{GraphMAE}~\citep{graphmae2022} is a GAE method where a subset of nodes in the input graph are masked. The masked graph is then input to an encoder (such as GCN~\citep{kipf2017} or GAT~\citep{petar2018gat}) to obtain node embeddings. In the decoding stage, the embeddings of the subset of nodes which were masked in the input graph are re-masked. The output of the decoder is used to reconstruct input node features of masked nodes, with a variant of cosine distance as a measure of loss. Extending this framework, \textit{GraphMAE2}~\citep{hou2023graphmae2} enhances feature reconstruction by introducing multi-view random re-masking and latent representation prediction. The random re-masking introduces stochasticity, preventing overfitting, while latent prediction leverages embedding space targets for masked nodes, making the model more robust to noise. GraphMAE and GraphMAE2’s focus on reconstructing the features of nodes masked in the input graph led to its superior performance on node classification tasks on benchmark datasets compared to contrastive methods. The authors also show that GraphMAE performs well in graph classification and inductive learning tasks. Moreover, \textit{GiGaMAE}~\citep{shi2023gigamae} achieved better results by shifting the reconstruction goal from the feature towards the latent space.

\textit{GCMAE}~\citep{wang2024generative}, which unifies both constrastive and masked auto-encoder techniques, is closest to the architecture we propose in this paper. More recently, \textit{MCGMAE}~\citep{fu2024multilevel} proposed a similar approach. However, they reported their model performance on only a few datasets. Both methods, \textit{GCMAE} and \textit{MCGMAE}, build upon the GraphMAE paradigm with additionally a dot product decoder to reconstruct the edges. \textit{MCGMAE} employs inter- and intra-class contrastive learning by generating positive and negative node pairs based on feature similarity, thereby reinforcing both node-level and class-level distinctions. In contrast, \textit{GCMAE} forms positive pairs by deriving node embeddings from two augmented views, one generated by node dropping and the other by feature masking, capturing both local and global structural information within the graph. By integrating these contrastive strategies with generative reconstruction, both \textit{MCGMAE} and \textit{GCMAE} achieve robust graph representations, demonstrating enhanced performance across tasks such as node classification and clustering.

We extend the success of these methods and present a novel architecture for graph SSL that integrates both contrastive and generative paradigms while addressing key limitations of existing approaches. Unlike \textit{GCMAE}, which primarily employs view-based node pair sampling, and \textit{MCGMAE}, which focuses on inter- and intra-class node sampling, our method introduces community-aware contrastive learning to capture structural information beyond local node neighborhoods or feature similarity. Additionally, we incorporate graph-level contrastive learning to enhance global semantic representation and ensure consistency across augmented graph views. To further improve robustness, we design a comprehensive augmentation strategy that jointly applies feature masking, node perturbation, and edge perturbation. These innovations allow our model to outperform state-of-the-art methods across multiple graph learning tasks, including node classification, clustering, and link prediction. The detailed contributions of our method are outlined below.

\subsection{Our Contributions}
\label{contributions}
\begin{figure}
\centering
\includegraphics[width=3.3in, height=2in, keepaspectratio]{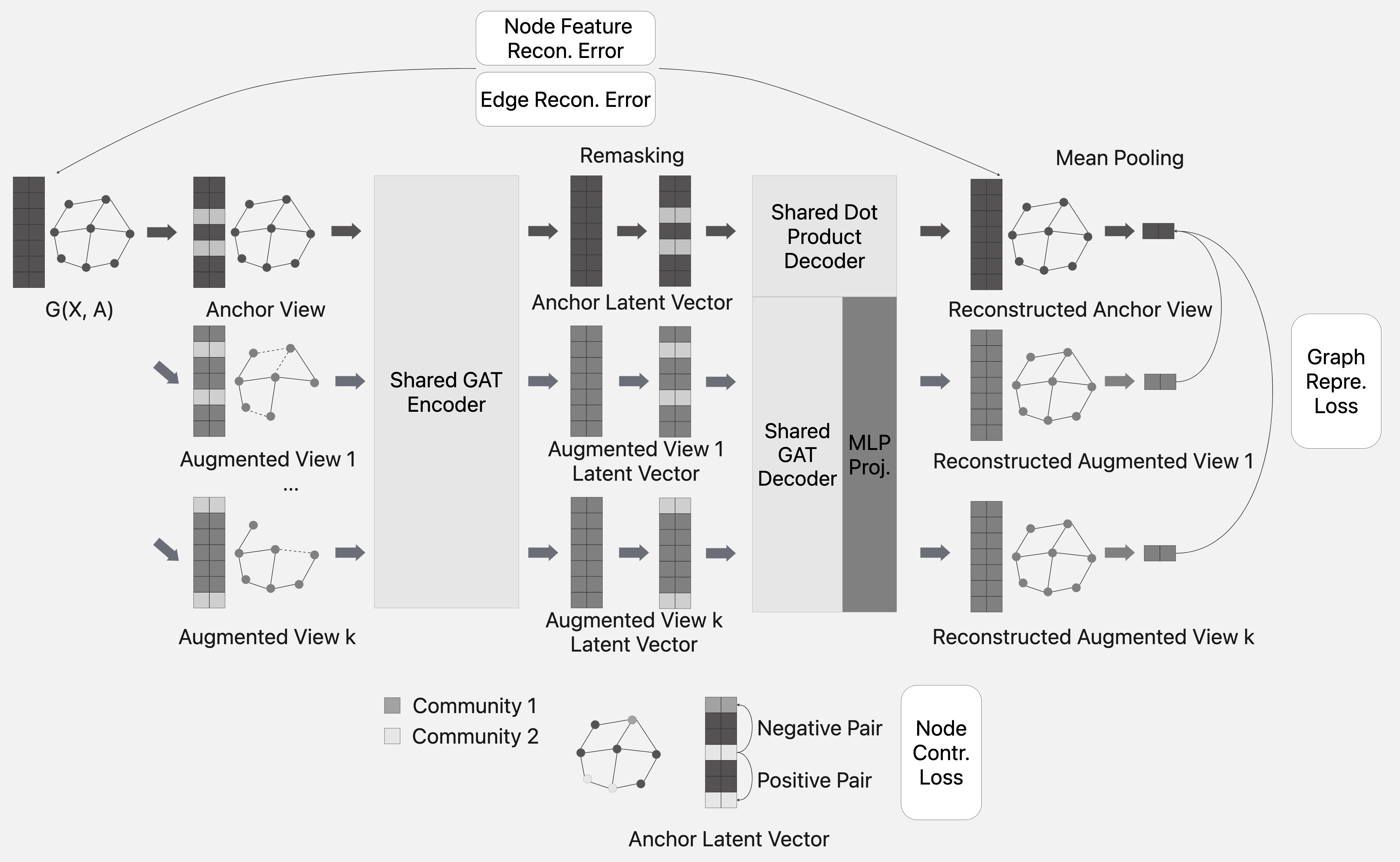}
\caption{Illustration of GCGRL Architecture. Given an input graph $\mathcal{G}=(\mathcal{V},\mathbf{A},\mathbf{X})$, the model generates an anchor view through node feature masking and $k$ augmented views (orange) through graph augmentation. A shared GAT encoder processes these views to obtain latent vectors, which undergo re-masking before decoder processing. The architecture employs parallel decoders: a shared dot product decoder for edge reconstruction and a shared GAT decoder with MLP projector for feature reconstruction. The model incorporates four loss functions: (1) node feature reconstruction error computed on the anchor view, (2) edge reconstruction error comparing decoded and original graph structures, (3) node contrastive loss using community-aware positive/negative pairs (whether the pair of nodes are coming from same community identified using the Louvain algorithm~\citep{louvain2008}) from the anchor latent vector, and (4) graph representation loss calculated between mean-pooled representations of anchor and augmented views.}
\label{fig:gcgrl_arch}
\end{figure}

To the best of our knowledge, we are the first to combine masked auto-encoder, node and graph level contrastive learning, community-aware node contrastive learning, and comprehensive augmentation strategies such as node perturbation, edge perturbation, and node feature masking together in one method. Our method performs well in all three tasks: node classification, node clustering, and link prediction. While Figure~\ref{fig:gcgrl_arch} illustrates the overall architecture of our model, the key novelties we bring are as follows:
\begin{itemize}
    \item \textbf{Community-Aware Contrastive Learning:} We introduce a novel community-driven node sampling strategy that leverages the Louvain algorithm~\cite{louvain2008} for positive/negative pair generation. Specifically, nodes within the same detected community are treated as positive pairs, while nodes from different communities serve as negative pairs. Compared to multi-view node pairing strategies used in \textit{MVGRL}~\citep{mvgrl2020} and \textit{GCMAE}~\citep{wang2024generative}, which rely on augmented views of the same node, our method captures higher-order structural relationships beyond local connectivity. As a result, community-based node pairs remain more stable under minor graph perturbations, enhancing training robustness.

    \item \textbf{Hybrid Contrastive Learning Framework:} We propose a novel architecture that operates at multiple levels simultaneously. At the node level, our community-guided contrastive learning preserves fine-grained structural patterns. At the graph level, we learn to capture global semantic information through masked reconstruction of the graph structure and contrastive learning at the graph view level. This multi-level approach enables our model to learn both local and global graph properties effectively, significantly improving performance across different downstream tasks.

    \item \textbf{Comprehensive Augmentation Strategy:} Unlike other contrastive learning methods, such as \textit{MVGRL}\citep{mvgrl2020}, \textit{BGRL}\citep{bgrl2022}, and \textit{GCMAE}~\citep{wang2024generative}, our approach employs a comprehensive augmentation strategy that integrates feature masking, node perturbation, and edge perturbation within a single view. This holistic approach allows the model to learn more robust and diverse representations by leveraging information from multiple perspectives.
\end{itemize}

These contributions introduce a robust framework that combines community awareness, hybrid-level learning, and the comprehensive augmentation strategy while addressing the limitations of the existing methods, particularly in how they handle structural information and generate contrastive signals.

\begin{figure}
\centering
\includegraphics[width=3.3in, height=2in, keepaspectratio]{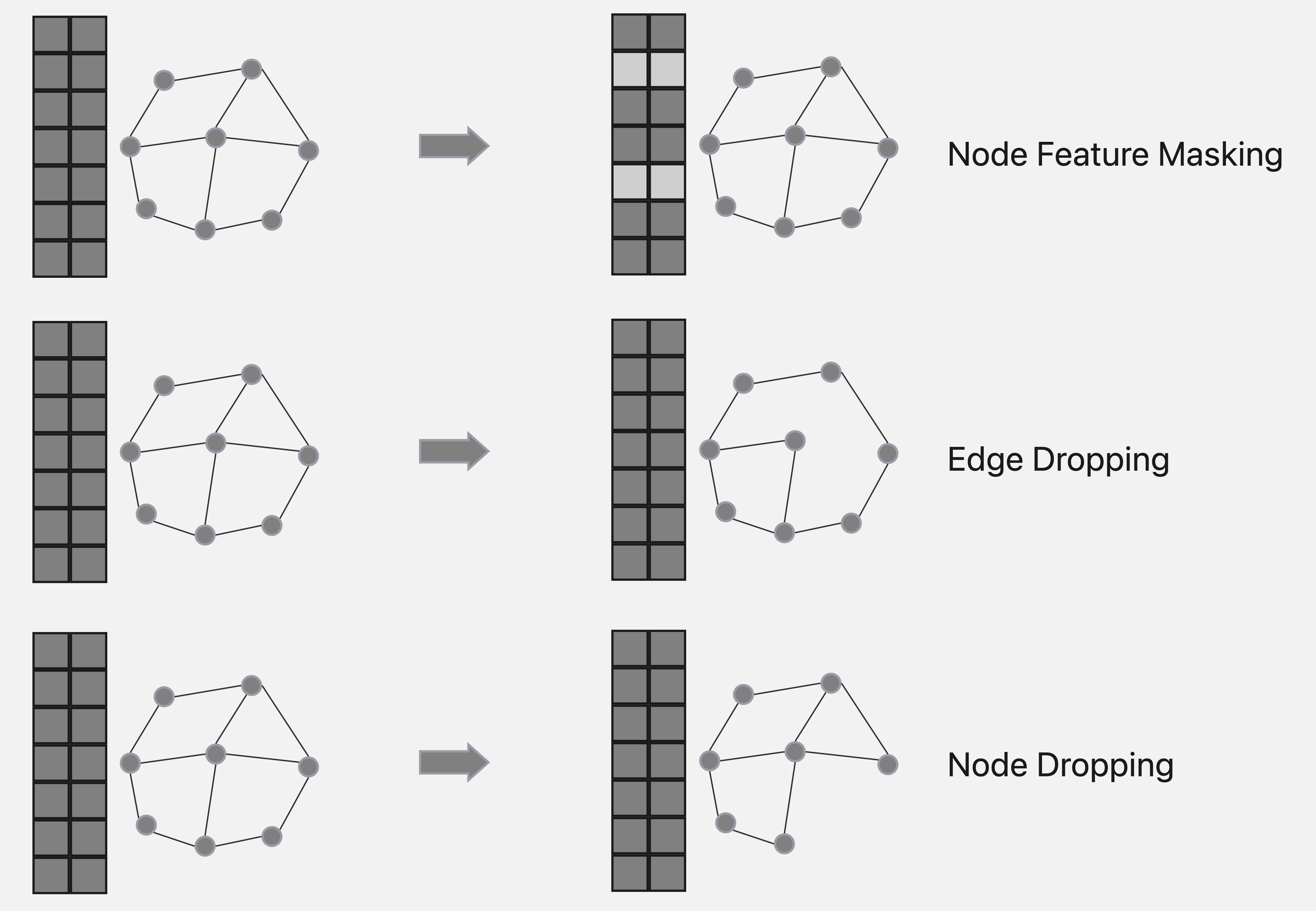}
\caption{Illustration of graph augmentation. Given an input graph $\mathcal{G}=(\mathcal{V},\mathbf{A},\mathbf{X})$, we apply a mixture of the three types of augmentations: (1) node feature masking, where features of randomly selected nodes are replaced with a learnable masking vector, (2) edge dropping, where a portion of edges are randomly removed while maintaining graph connectivity, and (3) node dropping, where selected nodes and their associated edges are removed from the graph. These augmentation strategies create diverse views of the input graph while preserving its essential structural and semantic information, enabling effective contrastive learning between anchor and augmented views.}
\label{fig:graph_aug}
\end{figure}


\section{GCGRL: Generative and Contrastive Graph Representation Learning}
\label{gcgrl}
We now elaborate on the components of our model GCGRL as illustrated in Figure~\ref{fig:gcgrl_arch}. The input graph is given by $\mathcal{G}=(\mathcal{V},\mathbf{A},\mathbf{X})$, where $\mathcal{V}$ is the set of nodes, $\mathbf{A}$ is the adjacency matrix, and $\mathbf{X}$ is the node feature matrix $\mathbb{R}^{|\mathcal{V}|\times d}$.

\subsection{Model Architecture} We first create an \textit{anchor} view from the input graph using node feature masking. More specifically, the feature values of a subset of nodes, $\widetilde{\mathcal{V}}\subset\mathcal{V}$ are set to a learnable vector $\mathbf{x}_{[M]}\in \mathbb{R}^{d}$. In parallel, a set of \textit{augmented} views of the input graph via a comprehensive mixture of node feature masking, node dropping, and edge dropping techniques as described in~\citep{you2020,zhu2020b,bgrl2022} are generated. Next, the anchor view $\mathcal{G}_{anchor}$ and the augmented views, $\mathcal{G}_{aug,i \in 1 \dots k}$, are input to a shared graph encoder $f_E$, a GAT~\citep{petar2018gat}, or a GCN~\citep{kipf2017} depends on datasets and experiments, in our model, to obtain node embeddings. The output of the encoder is a set of $k+1$ tensors, each of which represents the node embeddings of the anchor view and augmented views of the graph. We denote these hidden node representations as $\mathbf{H_{i \in 1 \dots k+1}} \in \mathbb{R}^{|\mathcal{V}|\times d_h}$, where $\mathbf{H_{1}}$ denotes the node embeddings of the anchor view, while $\mathbf{H_{i \in 2 \dots k+1}}$ are those of the augmented views. Next, for each of the hidden representations, $\mathbf{H_{i \in 1 \dots k+1}}$, we re-mask the set of nodes that were masked prior to the input to encoder. The remasking strategy is similar to what was described in GraphMAE~\citep{graphmae2022}. More specifically, we set the values of $v_i\in\widetilde{\mathcal{V}}$ in the hidden representation of $\mathcal{G}_{anchor}$ to a learnable vector $\mathbf{h}_{[M]}\in \mathbb{R}^{d_h}$. This is followed by two shared decoders that generate output node embeddings used for loss computations. We describe the decoders, how the contrastive pairs of graph views and nodes are generated, as well as loss computations below.

\subsection{Loss Functions}
Our model employs four complementary loss functions to capture different aspects of graph representation learning. The node contrastive loss leverages community structure detected by the Louvain algorithm~\citep{louvain2008} to identify meaningful positive and negative pairs, enhancing structural information learning. The node feature reconstruction loss ensures accurate recovery of masked node features, while the graph representation loss promotes consistency between anchor and augmented views at the graph level. Finally, the edge reconstruction loss preserves the graph's topological structure. These loss functions work in concert to learn comprehensive graph representations that are effective for various downstream tasks. We detail each loss function below.

\paragraph{Node Contrastive Loss} For a subset of nodes $\mathcal{V}_{\rho}$ of the anchor view $\mathcal{G}_{anchor}$, we obtain pairs of postive and negative matches. Positive pairs are nodes that belong to the same community generated by applying Louvain algorithm~\citep{louvain2008}, while negative pairs are nodes belonging to different communities. The corresponding node embeddings from the hidden layer, $\mathbf{H_{1}}$, are used to compute the contrastive loss function given by Eq.~\ref{eq:node_cl}:
\begin{equation}
\label{eq:node_cl}
\mathcal{L}_{nodeCL} = \frac{1}{2|\mathcal{V}_{\rho}|}\sum_{u \in \mathcal{V}_{\rho}}-\log{\frac{\sum\limits_{v \in V_u^+}e^{s(u,v)}}{\sum
\limits_{v \in V_u^+}e^{s(u,v)}+\sum\limits_{v \in V_u^-}e^{s(u,v)}}},
\end{equation}
where $\mathnormal{V}_{u^+}$ and $\mathnormal{V}_{u^-}$ denote the positive and negative matches, and $\mathnormal{s(.)}$ is the cosine similarity between embeddings of two nodes. Note that the node contrastive loss is computed using the node embeddings of $\mathcal{G}_{anchor}$ generated from the encoder and not from the output of the decoder.

\paragraph{Node Feature Reconstruction Loss} The hidden representation of the anchor view $\mathcal{G}_{anchor}$, $\mathbf{H_{1}}$, is passed through a decoder, $f_{D_1}$, which is a GAT~\citep{petar2018gat} followed by a MLP layer as illustrated in Figure~\ref{fig:gcgrl_arch}. Following the approach in GraphMAE~\citep{graphmae2022}, embeddings of masked nodes $\widetilde{\mathcal{V}}$ are remasked before input to the decoder. $f_{D_1}$ generates reconstructed node features, $\mathbf{Z} \in \mathbb{R}^{|\mathcal{V}|\times d}$, which are then used to compute node feature reconstruction loss using scaled cosine error (SCE)~\citep{graphmae2022} as given by Eq.~\ref{eq:node_sce}:
\begin{equation}
\label{eq:node_sce}
\mathcal{L}_{nodeFRL} = \frac{1}{|\widetilde{\mathcal{V}}|}\sum_{v_i \in \widetilde{\mathcal{V}}} (1 - \frac{\mathbf{x}^T_i \mathbf{z}_i}{\| \mathbf{x}_i\| \cdot \| \mathbf{z}_i\|})^{\gamma_1},~ \gamma_1 \ge 1,
\end{equation}
where $\mathbf{x}_i \in \mathbf{X}$ are the node features in the input graph $\mathcal{G}$, and $\gamma_1$ is a learnable parameter.

\paragraph{Graph Representation Loss} The decoder $f_{D_1}$ is also used to obtain output node embeddings of each of the augmented views $\mathcal{G}_{aug,i \in 1 \dots k}$. We then apply mean pooling to obtain graph representations of the anchor view as well as augmented views.  The graph representation loss, computed as the SCE between representations of the anchor view with respect to the augmented views, is given by Eq~\ref{eq:graph_rl}:
\begin{equation}
\label{eq:graph_rl}
\mathcal{L}_{graphRL} = \frac{1}{k}\sum_{i=2}^{k+1}(1 - \frac{\mathbf{w}^T_i \mathbf{w}_1}{\| \mathbf{w}_i\| \cdot \| \mathbf{w}_1\|})^{\gamma_2},~ \gamma_2 \ge 1,
\end{equation}
where $\mathbf{w_{i} \in \mathbb{R}^{d_g}}, i \in 1 \dots k+1$, denote the graph embeddings. The size of the embeddings is given by $d_g$; $\mathbf{w_1}$ is the graph embedding of the anchor view, and $\mathbf{w_{i \in 2 \dots k+1}}$ are those of the augmented views.

\paragraph{Edge Reconstruction Loss} The hidden representation of the anchor view $\mathcal{G}_{anchor}$, following remasking, is passed as input to another decoder $f_{D_2}$. We chose an inner product decoder~\citep{kipf2016} for this transformation. The output of $f_{D_2}$ is a probabilistic adjacency matrix $\mathbf{\widetilde{A}}$, which is then used in Eq~\ref{eq:edge_rl} to measure edge reconstruction loss. $\mathcal{E}^+$ is the set of postive edges, which are essentially the edges in the input graph, and $\mathcal{E}^-$ are the non-existent edges obtained via negative sampling. This loss function is similar to that used in MaskGAE~\citep{maskgae2023}.
\begin{equation}
\label{eq:edge_rl}
\mathcal{L}_{edgeRL} = -\biggl[\frac{1}{|\mathcal{E}^+|}\sum_{(u,v) \in \mathcal{E}^+}\log\mathbf{\widetilde{A}_{u,v}}+\frac{1}{|\mathcal{E}^-|}\sum_{(u,v) \in \mathcal{E}^-}\log(1-\mathbf{\widetilde{A}_{u,v}})\biggr]
\end{equation}

The overall loss is computed as the weighted sum of the above four loss functions as show in Eq~\ref{eq:loss}:
\begin{equation}
\label{eq:loss}
\mathcal{L} = \mathcal{L}_{nodeFRL}+\alpha_{1}\mathcal{L}_{nodeCL}+\alpha_{2}\mathcal{L}_{graphRL}+\alpha_{3}\mathcal{L}_{edgeRL},
\end{equation}
where $\alpha_{1}, \alpha_{2}, \alpha_{3}$ are tunable hyperparameters.

\section{Experimental Results}
\subsection{Model Performance on Benchmark Datasets}
\label{experimental_results}
\begin{table}
\caption{Benchmark datasets used in experiments.}
\label{tab:datasets}
\centering
\small
\begin{tabular}{lrrrrr}
\hline
& \textbf{Nodes} & \textbf{Edges} & \textbf{Features} & \textbf{Classes} \\
\hline
Cora  & 2,708 & 10,556 & 1,433 & 7\\
Citeseer  & 3,327 & 9,228 & 3,703 & 6\\
PubMed & 19,717	& 88,651 & 500 & 3 \\
Computers  & 13,752 & 245,861 & 767 & 10\\
Photos  & 7,650 & 119,081 & 745 & 8\\
CS & 18,333 & 81,894 & 6,805 & 15\\
Physics & 34,493 & 247,962 & 8,415 & 5\\
\hline
\end{tabular}
\end{table}

We run experiments to compare performance of GCGRL with existing models, described in Section~\ref{litreview}, on seven benchmark datasets. The dataset size varies from a few thousand nodes to hundreds of thousands of nodes and edges. The datasets and some of their metadata are described in Table~\ref{tab:datasets}. We compare the performance of our model on node classification, node clustering, and link prediction tasks.

\paragraph{Node Classification} For node classification tasks we follow the protocol applied by S2GAE~\cite{s2gae2023}. Average results with 95\% margin of error are reported from 10 runs of pre-training of GCGRL followed by linear SVC for classification afterwards. We measure the models' performance using a classification accuracy metric. As shown in Table~\ref{tab:nodecls}, GCGRL, outperforms the existing models (Section~\ref{litreview}) in the node classification task on five out of seven benchmark datasets. For the remaining two datasets, Amazon Photos and Coauthor CS, the performance of GCGRL is within the error bounds of the best model. Interestingly, GCGRL outperforms the constrastive methods BGRL and MVGRL in node classification tasks. This level of performance is an outcome of the constrastive loss function that attempts to minimize the feature reconstruction loss of masked nodes as described in Section~\ref{gcgrl}.

\begin{table}
\caption{Node classification accuracy.\protect\footnotemark}
\label{tab:nodecls}
\centering
\small
\renewcommand{\arraystretch}{1.2}
\resizebox{0.46\textwidth}{!}{
    \begin{tabular}{m{0.6in}|m{0.3in}m{0.3in}m{0.3in}m{0.3in}m{0.3in}m{0.3in}m{0.3in}m{0.3in}}
    \hline
    & Cora & Citeseer & PubMed & Com-puters & Photos & CS & Physics\\
    \hline
    MVGRL  & 85.86 ± 0.15 & 73.18 ± 0.22 & 84.86 ± 0.31 & 88.70 ± 0.24 & 92.15 ± 0.20 & 92.87 ± 0.13 & 95.35 ± 0.08 \\
    BGRL  & 86.16 ± 0.20 & 73.96 ± 0.14 & 86.42 ± 0.18 & 90.48 ± 0.10 & 93.22 ± 0.15 & 93.35 ± 0.06 & 96.16 ± 0.09 \\
    GraphMAE & 85.45 ± 0.40 & 72.48 ± 0.77 & 85.74 ± 0.14 & 88.04 ± 0.61 & 92.73 ± 0.17 & \textbf{93.47 ± 0.04} & 96.13 ± 0.03 \\
    MaskGAE*  & 87.31 ± 0.05 & 75.20 ± 0.07 & 86.56 ± 0.26 & 90.52 ± 0.04 & 93.33 ± 0.14 & 92.31 ± 0.05 & 95.79 ± 0.02 \\
    S2GAE  & 86.15 ± 0.25 & 74.60 ± 0.06 & 86.91 ± 0.28 & 90.94 ± 0.08 & 93.61 ± 0.10 & 91.70 ± 0.08 & 95.82 ± 0.03 \\
    GCMAE* & 86.88 ± 0.17 & 76.66 ± 0.35 & 82.67 ± 0.11 & 89.88 ± 0.06 & 92.76 ± 0.12 & 93.37 ± 0.08 & 96.24 ± 0.05 \\
    GiGaMAE* & 87.55 ± 0.19 & 73.71 ± 0.23 & 85.78 ± 0.04 & 90.59 ± 0.14 & \textbf{93.75 ± 0.06} & 93.16 ± 0.05 & 96.23 ± 0.02 \\
    \hline
    GCGRL & \textbf{87.78 ± 0.35} & \textbf{77.03 ± 0.34} & \textbf{88.70 ± 0.17} & \textbf{91.31 ± 0.08} & 93.59 ± 0.08 & 93.45 ± 0.06 & \textbf{96.52 ± 0.05} \\
    \hline
    \end{tabular}}
\end{table}

\footnotetext{We are citing experiment results from S2GAE~\cite{s2gae2023} except for GCMAE, MaskGAE and GiGaGAE where we run the experiments by ourselves, as marked by *}

\paragraph{Node Clustering} We apply K-Means clustering, with K being the number of classes for a given dataset as given in Table~\ref{tab:datasets}, using the node embeddings generated by GCGRL. In these experiments, the model is applied on the whole graph without any train/eval/test split. The models' performance for node clustering tasks is measured using two metrics: Normalized Mutual Information (NMI) and Adjusted Rand Index (ARI). The closer these metrics are to 1 the better the model's performance. As shown in Table~\ref{tab:nodeclus}, our model outperforms other models on the node clustering task when applied to Amazon-Computers, Photos, Coauthor-CS, and Physics, while it underperforms in small-mid size datasets such as Cora, CiteSeer and PubMed, though within the margin of error for PubMed.

\begin{table}
\caption{Node clustering results.\protect\footnotemark}
\label{tab:nodeclus}
\centering
\small
\renewcommand{\arraystretch}{1.2}
\resizebox{0.46\textwidth}{!}{
    \begin{tabular}{m{0.4in}m{0.3in}|m{0.25in}m{0.25in}m{0.25in}m{0.25in}m{0.25in}m{0.25in}m{0.25in}m{0.25in}}
    \hline
    & Metric & Cora & Citeseer & PubMed* & Com-puters & Photos & CS & Physics \\
    \hline
    \multirow{2}{0.7in}{MVGRL}
    & NMI & 0.5481 & 0.4073 & 0.3075 & 0.2657 & 0.1776 & 0.7736 & 0.4948 \\
    & ARI & 0.5167 & 0.4115 & 0.3042 & 0.1806 & 0.1127 & 0.6646 & 0.4799 \\
    \multirow{2}{0.7in}{BGRL}
    & NMI & 0.2851 & 0.2156 & 0.2173 & 0.4396 & 0.6189 & 0.7740 & 0.7249 \\
    & ARI & 0.0920 & 0.1759 & 0.1725 & 0.2096 & 0.4754 & 0.6422 & 0.8130 \\
    \multirow{2}{0.7in}{GraphMAE}
    & NMI & 0.5821 & 0.4330 & \textbf{0.3459} & 0.5015 & 0.6676 & 0.7297 & 0.6348 \\
    & ARI & 0.5310 & 0.4423 & \textbf{0.3325} & 0.3298 & 0.5703 & 0.5691 & 0.6734 \\
    \multirow{2}{0.7in}{MaskGAE*}
    & NMI & 0.4708 & 0.3716 & 0.3299 & 0.5178 & 0.6456 & 0.7362 & 0.6056 \\
    & ARI & 0.4220 & 0.3690 & 0.3168 & 0.3534 & 0.5404 & 0.5899 & 0.4776 \\
    \multirow{2}{0.7in}{S2GAE}
    & NMI & 0.5127 & 0.3346 & 0.3148 & 0.4397 & 0.5624 & 0.6251 & 0.6152 \\
    & ARI & 0.4481 & 0.2830 & 0.3086 & 0.2297 & 0.3427 & 0.4289 & 0.7059 \\
    \multirow{2}{0.7in}{GCMAE*}
    & NMI & 0.3563 & \textbf{0.4453} & 0.3442 & 0.4359 & 0.5205 & 0.5877 & 0.6268 \\
    & ARI & 0.2413 & \textbf{0.4555} & 0.3275 & 0.2453 & 0.3362 & 0.4263 & 0.4427 \\
    \multirow{2}{0.7in}{GiGaMAE}
    & NMI & \textbf{0.5836} & 0.4224 & 0.2837 & 0.5228 & 0.7066 & 0.7622 & 0.7373 \\
    & ARI & \textbf{0.5453} & 0.4283 & 0.2641 & 0.3579 & 0.5859 & 0.6417 & 0.8271 \\
    \hline
    \multirow{2}{0.7in}{GCGRL}
    & NMI & 0.5821 & 0.4268 & 0.3450 & \textbf{0.5471} & \textbf{0.7142} & \textbf{0.7690} & \textbf{0.7606} \\
    & ARI & 0.5350 & 0.4422 & 0.3241 & \textbf{0.4222} & \textbf{0.6180} & \textbf{0.6838} & \textbf{0.8317} \\
    \hline
    \end{tabular}}
\end{table}

\footnotetext{We are citing experiment results from GiGaMAE~\cite{shi2023gigamae} except for PubMed dataset, GCMAE and MaskGAE where we run the experiments by ourselves, as marked by *.}

\paragraph{Link Prediction} For link prediction tasks we are following the setting of GiGaMAE~\cite{shi2023gigamae} to measure the models' performance using the area under ROC curve (AUC). The edges are split into 85\% training, 10\% testing, and 5\% validation sets. We only use training edges for model pretraining and validation edges for hyperparameter tuning. Edge reconstruction error component in the loss function results in impressive performance on link prediction tasks. As shown in Table~\ref{tab:linkpred} the GCGRL model outperforms the existing contrastive and generative models on five out of seven datasets. In the case of PubMed and Computers data, GCGRL's performance is very close to the best model. Also, unsurprisingly, GAE methods, by virtue of edge reconstruction error loss as an objective, outperform contrastive methods such as BGRL and MVGRL on link prediction tasks by a wide margin. We note that GraphMAE, unlike other GAE methods, does not perform well on link prediction tasks as it does not try to minimize edge reconstruction error.

\begin{table}
\caption{Link prediction results.\protect\footnotemark}
\label{tab:linkpred}
\centering
\small
\renewcommand{\arraystretch}{1.2}
\resizebox{0.46\textwidth}{!}{
    \begin{tabular}{m{0.6in}|m{0.3in}m{0.3in}m{0.3in}m{0.3in}m{0.3in}m{0.3in}m{0.3in}m{0.3in}}
    \hline
    & Cora & Citeseer & PubMed* & Com-puters & Photos & CS & Physics \\
    \hline
    MVGRL & 74.57 ± 0.38 & 68.33 ± 0.59 & 88.36 ± 0.59 & 85.32 ± 0.25 & 84.89 ± 0.08 & 77.13 ± 0.33 & 77.26 ± 0.53 \\
    BGRL & 91.70 ± 0.59 & 92.90 ± 0.57 & 96.75 ± 0.12 & 93.69 ± 0.43 & 94.21 ± 0.64 & 92.60 ± 0.15 & 92.29 ± 0.56 \\
    GraphMAE & 88.78 ± 0.87 & 90.32 ± 1.26 & 93.72 ± 0.00 & 74.04 ± 3.08 & 74.58 ± 3.90 & 85.37 ± 1.37 & 81.29 ± 5.13 \\
    MaskGAE* & 96.45 ± 0.18 & 98.02 ± 0.22 & \textbf{98.84 ± 0.04} & \textbf{97.98 ± 0.05} & 98.62 ± 0.06 & 97.00 ± 0.10 & 97.35 ± 0.07 \\
    S2GAE & 93.12 ± 0.58 & 93.81 ± 0.23 & 98.45 ± 0.03 & 94.59 ± 1.16 & 93.84 ± 2.22 & 96.13 ± 0.48 & 95.21 ± 0.75 \\
    GCMAE* \footnotemark & 93.43 ± 0.33 & 93.53 ± 0.36 & 96.93 ± 0.10 & 93.52 ± 0.10 & 95.37 ± 0.07 & 97.13 ± 0.11 & 94.66 ± 0.01 \\
    GiGaMAE & 95.13 ± 0.15 & 94.18 ± 0.36 & 93.67 ± 0.19 & 95.17 ± 0.39 & 96.24 ± 0.11 & 96.34 ± 0.07 & 96.32 ± 0.08 \\
    \hline
    GCGRL & \textbf{98.46 ± 0.21} & \textbf{98.98 ± 0.13} & 98.33 ± 0.10 & 96.93 ± 0.14 & \textbf{98.77 ± 0.04} & \textbf{99.12 ± 0.05} & \textbf{98.54 ± 0.03} \\
    \hline
    \end{tabular}}
\end{table}

\addtocounter{footnote}{-1}
\footnotetext{We are citing experiment results from GiGaMAE~\cite{shi2023gigamae} except for the PubMed dataset, GCMAE and MaskGAE where we run the experiments by ourselves, as marked by *.}

\addtocounter{footnote}{1}
\footnotetext{We've noticed implementation bugs for link prediction experiments in GCMAE, including:
    1.The testing edges are also included in the training process;
    2.The edge reconstruction loss is incorrectly defined as
$\mathcal{L}(\sigma(\mathbf{\widetilde{A}}_{u^+,v^+}), \sigma(\mathbf{\widetilde{A}}_{u^-,v^-}))$, where $\mathcal{L}$ is the binary cross entropy function, which is teaching the model to predict the same for positive and negative edges. The results are reproduced after the bugs are fixed.
}

\subsection{Ablation Studies}
We conduct ablation studies to validate the contributions of key design choices in GCGRL. For evaluation, we follow the same evaluation method as mentioned in Section~\ref{experimental_results} to assess the performance impact of each component.

\paragraph{Key Components}
To evaluate the contributions of each key component in our model, we conduct an ablation study across three core tasks: node classification, link prediction, and node clustering. Starting with a baseline configuration—a GNN-based Graph Autoencoder (GAE) that employs only masking and re-masking techniques, we incrementally add each of the key components: community-aware node-wise contrastive learning, graph-wise contrastive learning, and edge reconstruction on Cora, CiteSeer, Computers and Photo. This progressive inclusion allows us to isolate and observe the specific impact of each component on model performance for each task. By measuring the effects of these additions across tasks, we gain insights into how these components collectively enhance the model's representation learning and predictive accuracy.

\begin{table}[h]
    \centering
    \caption{Ablation Study for Key Components - Node Classification}
    \label{tab:ablation_node_cls}
    \small
    \renewcommand{\arraystretch}{1.2}
    \resizebox{0.46\textwidth}{!}{
    \begin{tabular}{m{0.6in}|m{0.4in}|m{0.4in}m{0.4in}m{0.4in}m{0.4in}}
        \hline
        Contr. & Metric & Cora & CiteSeer & Computers & Photo \\
        \hline
        None & ACC & 87.80 ± 0.31 & 77.11 ± 0.39 & 88.83 ± 0.69 & 89.63 ± 1.63 \\
        Node Contr. Only & ACC & 87.80 ± 0.30 & 77.11 ± 0.38 & 88.81 ± 0.70 & 91.07 ± 0.14 \\
        Both Contr. & ACC & 87.82 ± 0.32 & 77.10 ± 0.37 & 89.42 ± 0.58 & 92.57 ± 0.17 \\
        All & ACC & 87.78 ± 0.35 & 77.03 ± 0.34 & 91.31 ± 0.08 & 93.59 ± 0.08 \\
        \hline
    \end{tabular}}
\end{table}

\begin{table}[h]
    \centering
    \caption{Ablation Study for Key Components - Link Prediction}
    \label{tab:ablation_link_pred}
    \small
    \resizebox{0.46\textwidth}{!}{
    \renewcommand{\arraystretch}{1.2}
    \begin{tabular}{m{0.6in}|m{0.4in}|m{0.4in}m{0.4in}m{0.4in}m{0.4in}}
        \hline
        Contr. & Metric & Cora & CiteSeer & Computers & Photo \\
        \hline
        None & AUC & 84.82 ± 0.94 & 88.77 ± 0.99 & 42.79 ± 8.08 & 50.00 ± 0.00 \\
        Node Contr. Only & AUC & 94.77 ± 0.31 & 90.04 ± 0.87 & 78.81 ± 0.73 & 59.58 ± 10.92 \\
        Both Contr. & AUC & 94.72 ± 0.28 & 91.24 ± 0.95 & 79.17 ± 0.83 & 61.53 ± 11.57 \\
        All & AUC & 98.46 ± 0.21 & 98.98 ± 0.13 & 96.93 ± 0.14 & 98.78 ± 0.05 \\
        \hline
    \end{tabular}}
\end{table}

\begin{table}[h]
    \centering
    \caption{Ablation Study for Key Components - Node Clustering}
    \label{tab:ablation_node_clus}
    \small
    \renewcommand{\arraystretch}{1.2}
    \resizebox{0.46\textwidth}{!}{
    \begin{tabular}{m{0.6in}|m{0.4in}|m{0.4in}m{0.4in}m{0.4in}m{0.4in}}
        \hline
        Contr. & Metric & Cora & CiteSeer & Computers & Photo \\
        \hline
        \multirow{2}{0.6in}{None}
        & NMI & 0.4440 ± 0.0175 & 0.3923 ± 0.0160 & 0.4667 ± 0.0196 & 0.6017 ± 0.0555 \\
        & ARI & 0.3091 ± 0.0230 & 0.3822 ± 0.0262 & 0.3099 ± 0.0158 & 0.4687 ± 0.0532 \\
        \multirow{2}{0.6in}{Node Contr. Only}
        & NMI & 0.5239 ± 0.0135 & 0.3852 ± 0.0153 & 0.4714 ± 0.0131 & 0.6548 ± 0.0194 \\
        & ARI & 0.4181 ± 0.0282 & 0.3707 ± 0.0257 & 0.3761 ± 0.0262 & 0.5299 ± 0.0393 \\
        \multirow{2}{0.6in}{Both Contr.}
        & NMI & 0.5370 ± 0.0182 & 0.3949 ± 0.0187 & 0.4887 ± 0.0073 & 0.7095 ± 0.0219 \\
        & ARI & 0.4396 ± 0.0301 & 0.3854 ± 0.0318 & 0.3866 ± 0.0100 & 0.6085 ± 0.0187 \\
        \multirow{2}{0.6in}{All}
        & NMI & \textbf{0.5821 ± 0.0119} & \textbf{0.4268 ± 0.0066} & \textbf{0.5498 ± 0.0127} & \textbf{0.7142 ± 0.0247} \\
        & ARI & \textbf{0.5350 ± 0.0446} & \textbf{0.4422 ± 0.0092} & \textbf{0.4305 ± 0.0209} & \textbf{0.6180 ± 0.0494} \\
        \hline
    \end{tabular}}
\end{table}

\textit{Node Classification.}
As shown in Table~\ref{tab:ablation_node_cls}, while performance gains remain modest on the smaller datasets like Cora and CiteSeer, larger datasets exhibit substantial improvements with the full model. For Computers, node classification accuracy improves from 88.83\% to 91.31\%, whereas for the Photo dataset the increment is from 89.63\% to 93.59\%. The addition of both contrastive components leads to strong effects, especially on the Photo dataset, with progressive improvements of +1.44\% (Node Contrastive Only), +2.94\% (Both Contrastive), and +3.96\% (All components). These results highlight the effectiveness of integrating all components for node classification on larger datasets.

\textit{Link Prediction.}
Table~\ref{tab:ablation_link_pred} demonstrates that link prediction gains significantly from the proposed architecture. The baseline model struggles with the larger datasets Computers (42.79\%) and Photo (50.00\%) due to the increased structural complexity and sparsity of connections in these graphs. Node contrastive learning yields substantial initial improvements (+36.02\% on Computers, +9.58\% on Photo), enabling the model to capture discriminative node features effectively. The addition of graph-wise contrastive learning provides further boosts (+0.36\% on Computers, +1.95\% on Photo), while edge reconstruction explicitly preserves structural details, resulting in further improvements (+17.76\% on Computers, +11.78\% on Photo). The full model achieves consistently high performance across all datasets, underlining the importance of integrating contrastive learning and edge reconstruction for robust link prediction.

\textit{Node Clustering.}
Table~\ref{tab:ablation_node_clus} illustrates the impact of different components on node clustering performance, measured using NMI and ARI. The baseline model exhibits moderate clustering effectiveness, with NMI values of 0.4440 (Cora), 0.3923 (CiteSeer), 0.4667 (Computers), and 0.6017 (Photo). Introducing node-wise contrastive learning significantly enhances NMI and ARI across Cora, Computers and Photo, particularly in Cora (+0.0799 NMI, +0.1090 ARI) and Photo (+0.0531 NMI, +0.0612 ARI), demonstrating the benefits of enforcing community based structural learning. Further improvements emerge when incorporating graph-wise contrastive learning, which enhances global structural coherence and semantic understanding, leading to strong gains in the Photo dataset (NMI: +0.1078, ARI: +0.0786). The full model, integrating all components, achieves the highest NMI and ARI scores across all datasets, with particularly pronounced improvements in Cora (+0.0451 NMI, +0.0954 ARI) and Computers (+0.0831 NMI, +0.0544 ARI), emphasizing the synergy between node-wise contrastive learning, graph-wise contrastive learning, and edge reconstruction. These results confirm that a comprehensive contrastive and generative learning approach is essential for robust and high-quality node clustering.

These findings validate our architectural choices and demonstrate the complementary nature of contrastive and generative approaches in graph representation learning, with each component contributing uniquely to different aspects of model performance.

\begin{figure}[h]
    \centering
    \includegraphics[width=0.4\textwidth]{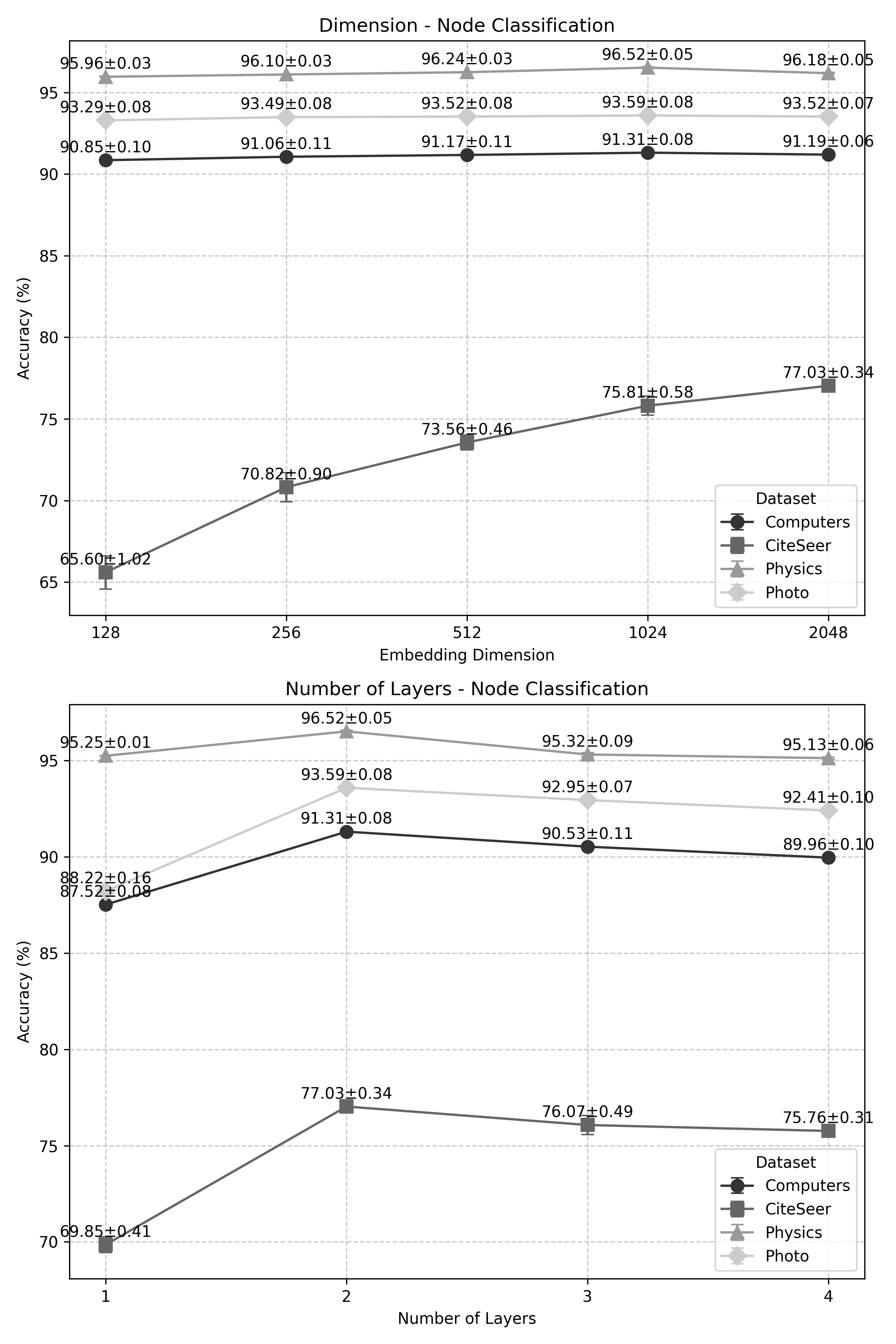}
    \caption{Model Size Ablation (Node Classification)}
    \label{fig:model_size}
\end{figure}

\paragraph{Model Size}To explore the impact of model size on node classification performance, we vary the embedding dimension and evaluate the results on different datasets. The performance metrics for various dimensions are shown in Figure \ref{fig:model_size}.

As shown in Figure \ref{fig:model_size}, increasing the embedding dimension generally improves classification accuracy across datasets. For the Physics dataset, accuracy increases from 95.96±0.03 at 128 dimensions to 96.52±0.05 at 1024 dimensions, with a decrease to 96.18±0.05 at 2048 dimensions. For the Photo and Computers datasets we observe similar patterns, while for CiteSeer we observe the most substantial improvement in node classification accuracy, which increases from 65.60±1.62 at 128 dimensions to 77.03±0.34 at 2048 dimensions. These results suggest that higher dimensional embeddings better capture complex data relationships; however, the benefits diminish beyond 1024 dimensions for larger datasets, possibly due to overfitting. For layer depth, model performance peaks at 2 layers across all datasets and thereafter drops when more layers are added.

These findings suggest that a configuration with 1024 dimensions and 2 layers provides optimal performance for most datasets in our framework. The degradation in deeper architectures indicates that simpler configurations often achieve better results, potentially due to the oversmoothing problem as discussed in ~\citep{rusch2023survey}.

\begin{table}[h]
    \centering
    \caption{Ablation Study for Model Type - Node Classification}
    \label{tab:model_type_classification}
    \resizebox{0.46\textwidth}{!}{
    \begin{tabular}{m{0.5in}|m{0.5in}|m{0.45in}m{0.45in}m{0.45in}m{0.45in}}
        \hline
        Encoder & Decoder & Cora & PubMed & Computers & Photo \\
        \hline
        MLP & MLP & 30.33 ± 0.04 & 81.98 ± 0.32 & 47.70 ± 0.18 & 67.57 ± 2.36 \\
        GCN & GCN & 85.00 ± 0.42 & \textbf{88.70 ± 0.17} & 88.54 ± 0.08 & 90.62 ± 0.09 \\
        GAT & GAT & \textbf{87.78 ± 0.35} & 86.69 ± 0.10 & \textbf{91.30 ± 0.09} & \textbf{93.59 ± 0.08} \\
        \hline
    \end{tabular}}
\end{table}

\begin{table}[h]
    \centering
    \caption{Ablation Study for Model Type - Link Prediction}
    \label{tab:model_type_link_prediction}
    \resizebox{0.46\textwidth}{!}{
    \begin{tabular}{m{0.5in}|m{0.5in}|m{0.45in}m{0.45in}m{0.45in}m{0.45in}}
        \hline
        Encoder & Decoder & Cora & PubMed & Computers & Photo \\
        \hline
        MLP & MLP & 73.47 ± 1.51 & 58.82 ± 1.06 & 78.91 ± 1.43 & 74.01 ± 0.76 \\
        GCN & GCN & \textbf{98.46 ± 0.21} & \textbf{98.33 ± 0.10} & \textbf{96.93 ± 0.14} & \textbf{98.77 ± 0.04} \\
        GAT & GAT & 97.94 ± 0.15 & 94.56 ± 0.19 & 95.80 ± 0.06 & 97.15 ± 0.14 \\
        \hline
    \end{tabular}}
\end{table}

\begin{table}[h]
    \centering
    \caption{Ablation Study for Model Type - Node Clustering}
    \label{tab:model_type_clustering}
    \resizebox{0.46\textwidth}{!}{
    \begin{tabular}{m{0.5in}|m{0.5in}|m{0.45in}m{0.45in}m{0.45in}m{0.45in}}
        \hline
        Encoder Decoder Type & Metrics & Cora & PubMed & Com-puters & Photo \\
        \hline
        \multirow{2}{*}{MLP} & NMI & 0.0392 ± 0.0183 & 0.1226 ± 0.0557 & 0.1915 ± 0.0169 & 0.2464 ± 0.0105\\
        & ARI & 0.0187 ± 0.0110 & 0.1121 ± 0.0589 & 0.0658 ± 0.0107 & 0.1292 ± 0.0106\\
        \multirow{2}{*}{GCN} & NMI & 0.5429 ± 0.0221 & 0.3047 ± 0.0795 & 0.5422 ± 0.0264 & 0.6999 ± 0.0126\\
        & ARI & 0.4805 ± 0.0401 & 0.2652 ± 0.0915 & 0.3710 ± 0.0391 & 0.6065 ± 0.0170 \\
        \multirow{2}{*}{GAT} & NMI & \textbf{0.5821 ± 0.0119} & \textbf{0.3450 ± 0.0018} & \textbf{0.5471 ± 0.0181} & \textbf{0.7142 ± 0.0247}\\
        & ARI & \textbf{0.5350 ± 0.0446} & \textbf{0.3241 ± 0.0021} & \textbf{0.4222 ± 0.0446} &  \textbf{0.6180 ± 0.0494}\\
        \hline
    \end{tabular}}
\end{table}

\paragraph{Model Type}
This experiment compares three encoder-decoder architectures (MLP, GCN, and GAT) across four different datasets, results are shown in the tables ~\ref{tab:model_type_classification}, ~\ref{tab:model_type_link_prediction} and ~\ref{tab:model_type_clustering}. This experiment demonstrates that graph-based architectures (GCN, GAT) significantly outperform MLP across all tasks. For node classification, GAT achieves the best performance on three datasets (Cora: 87.78\%, Computers: 91.30\%, Photo: 93.59\%), while GCN leads on PubMed (88.70\%). For link prediction tasks, GCN demonstrates superior performance across all datasets with the notably highest AUC scores. For node clustering tasks, GAT consistently outperforms other architectures. The consistent underperformance of MLP across all tasks underscores the importance of incorporating graph structural information and message passing in the learning process.

\section{Concluding Remarks}
We present GCGRL, a novel self-supervised graph learning framework that integrates both generative and contrastive learning paradigms. Unlike existing contrastive approaches that rely on augmented views of the same node, our method introduces a community-aware sampling strategy, leveraging the Louvain algorithm to construct semantically meaningful positive and negative pairs, enabling the model to capture higher-order structural relationships beyond local connectivity. In addition to node-level contrastive learning, our framework generates contrastive pairs at the graph level to capture global semantic information. Simultaneously, we incorporate generative objectives, including feature and edge reconstruction, to preserve fine-grained structural and topological information. Furthermore, we employ a comprehensive augmentation strategy that integrates feature masking, node perturbation, and edge perturbation, ensuring more diverse and expressive representations. We conduct extensive experiments on seven public datasets and demonstrate that by synergizing these features, GCGRL outperforms various state-of-the-art self-supervised graph learning models, demonstrating its effectiveness and robustness across node classification, node clustering, and link prediction.

Beyond strong empirical results, the ability of GCGRL to learn both global structural dependencies and fine-grained local relationships suggests potential applications in various graph-based tasks. In domains such as fraud detection and recommendation systems, where structural patterns and higher-order connectivity play a crucial role, the integration of community-aware contrastive learning and global contrastive objectives may offer advantages in capturing meaningful patterns. While this work does not explicitly evaluate these applications, future research may explore how GCGRL can be adapted to such scenarios. Further directions include extending GCGRL to heterogeneous, dynamic, and large-scale graphs, as well as optimizing its computational efficiency for real-world deployment. We believe that our findings provide new insights into the interplay between contrastive and generative paradigms in self-supervised learning, paving the way for more effective graph representation learning frameworks.

\bibliographystyle{ACM-Reference-Format}
\bibliography{reference}

\end{document}